# Differentiable Mobile Display Photometric Stereo

Gawoon Ban[1*]  Hyeongjun Kim[2*]  Seokjun Choi[1]  Seungwoo Yoon[1]

Seung-Hwan Baek[1]

POSTECH[1]     Sogang University[2]

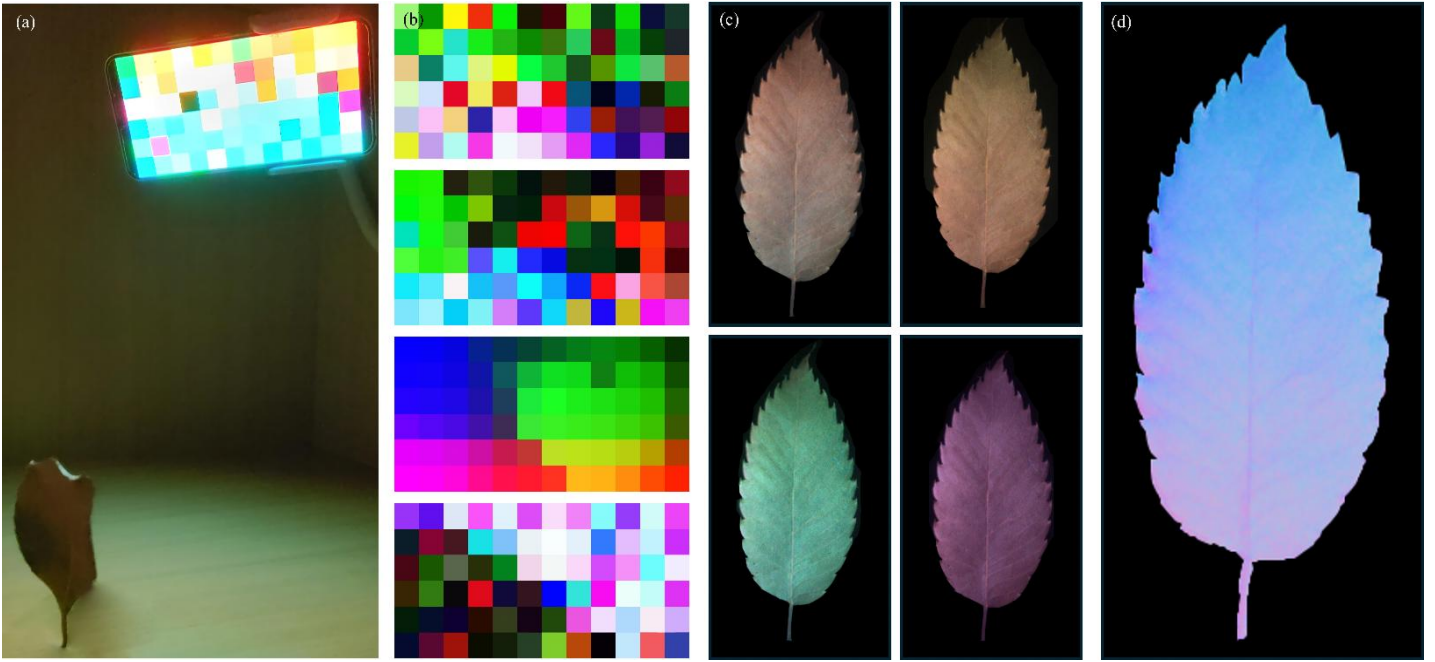

Figure 1. We propose differentiable mobile display photometric stereo (DMDPS), a method for high-quality photometric stereo using (a) a camera and display on a mobile phone. We display (b) learned patterns on the mobile display and capture (c) corresponding images using the mobile front-side camera. (d) Analyzing captured images, we demonstrate high-quality normal reconstruction.

## Abstract

Display photometric stereo uses a display as a programmable light source to illuminate a scene with diverse illumination conditions. Recently, differentiable display photometric stereo (DDPS) [1] demonstrated improved normal reconstruction accuracy by using learned display patterns. However, DDPS faced limitations in practicality, requiring a fixed desktop imaging setup using a polarization camera and a desktop-scale monitor. In this paper, we propose a more practical physics-based photometric stereo, differentiable mobile display photometric stereo (DMDPS), that leverages a mobile phone consisting of a display and a camera. We overcome the limitations of using a mobile device by developing a mobile app and method that simultaneously displays patterns and captures high-quality HDR images. Using this technique, we capture real-world 3D-printed objects and learn display patterns via a differentiable learning process. We demonstrate the effectiveness of DMDPS on both a 3D printed dataset and a first dataset of fallen leaves. The leaf dataset contains reconstructed surface normals and albedos of fallen leaves that may enable future research beyond computer graphics and vision. We believe that DMDPS takes a step forward for practical physics-based photometric stereo.

---

[*] Both of these contributed equally to the work.

## 1. Introduction

Estimating high-quality surface normals is a long-standing problem in computer vision and graphics. Display photometric stereo reconstructs surface normals using conventional monitors and cameras. Differentiable display photometric stereo (DDPS) enhances this process by learning illumination patterns, resulting in improved surface normal reconstruction compared to heuristic display patterns. However, DDPS has a limitation in that it uses a desktop environment with a fixed large LCD monitor and a polarization camera. The large LCD monitor is neither portable nor adaptable, requiring objects to be transported to the fixed setup, which limits flexibility.

This paper introduces Differentiable Mobile Display Photometric Stereo (DMDPS). We use a mobile phone as a portable device with a display and an on-device camera. This setup resolves the desktop fixed-location issue. However, using a phone as an illumination source reduces both the light-view angular samples and intensity of light. We address this challenge by increasing exposure time and using HDR imaging. Additionally, since most mobile phones do not allow simultaneous use of the light source and the camera, a custom mobile application was developed to overcome this limitation. The app enables precise control over ISO, exposure time, and frame duration, displays desired patterns, and captures objects using the front camera. It also supports RAW image capture and allows for multiple exposure times through preset configuration.

To enable image capture in general environments rather than dark rooms, HDR and RAW images are employed. External light in non-controlled settings can degrade the quality of surface normal reconstruction. To mitigate this, the system uses RAW images (without post-processing) and HDR images captured at varying exposure times. Furthermore, mobile phone cameras, which typically have lens-based image distortion (e.g., radial distortion), are corrected using intrinsic calibration and undistortion techniques, thereby improving accuracy.

In summary, our contributions are as follows:
· We propose Differentiable Mobile Display Photometric Stereo (DMDPS) that allows surface normal reconstruction using a mobile phone in general environments instead of a dark room and eliminates the need for monitors and polarized cameras.
· We developed a custom mobile app enabling RAW image capture, multiple exposure configurations, ISO value control, and more.
· We applied DMDPS to real-world objects (e.g., fallen leaves), and revealed surface normals and albedos reconstructed from these objects.

## 2. Related Work

**Display and Imaging Systems for Photometric Stereo** Various imaging systems for photometric stereo have been proposed, including DSLR camera flashes [3, 4], LCD monitors [1], and mobile phone flashes [5, 6]. Additionally, LCD monitors and polarization cameras [1] have been used to leverage diffuse images created by polarized light in display photometric stereo. In this paper, we perform photometric stereo using only the screen and camera of a mobile phone. Compared to the aforementioned imaging systems, this approach requires just one device—a mobile phone—making it the most portable solution. To implement DMDPS on mobile devices, we developed a custom app that supports RAW image capture, exposure time and ISO value control, and multi-exposure time capturing through presets.

**Learned Illumination Pattern** A critical challenge in photometric stereo is determining the optimal illumination pattern. The illumination pattern defines how the intensity of light sources is distributed, and selecting an effective pattern is crucial for accurate surface normal reconstruction. One commonly used standard is the one-light-at-a-time (OLAT) pattern, where each light source is activated at maximum intensity one at a time [7, 8]. This method is effective when each light source provides sufficient energy for the camera sensor to capture without introducing significant noise [9]. While several heuristic patterns exist, some methods employ learned patterns optimized for high-quality surface normal reconstruction. These patterns are generated by comparing estimated normal maps with ground truth data and creating error maps across different patterns [2]. In our work, we enhance the learning process by introducing a Gaussian filter and adjusting the learning rate. Because our imaging system transitions from using an LCD monitor and polarization camera to a mobile phone, the Gaussian filter's sigma value and the learning rate are fine-tuned to accommodate the new imaging environment.

**Capturing Environment** Conventional photometric stereo method typically capture images in a dark room [1, 10]. This is because external light sources can interfere with the designated illumination pattern, reducing its effectiveness and resulting in similar photographed images across different patterns. To address this limitation, we propose capturing images in general environments using HDR imaging. Instead of relying solely on standard RAW images, HDR images are captured at multiple exposure times to improve robustness against external light interference.

**Photometric Stereo Dataset** Various datasets have been proposed [12, 13, 14] for evaluating or training photometric stereo methods, including both synthetic [15] and real-world datasets. In our work, we utilize real-world datasets, obtaining ground truth for training datasets using 3D-printed objects. Additionally, we applied DMDPS to in-the-wild objects, such as fallen leaves, and revealed the reconstructed surface normals and albedos through this process.

## 3. DDPS vs DMDPS

**3.1. DDPS Review** DDPS focuses on designing and learning illumination patterns to achieve accurate surface normal reconstruction of objects. The process is broadly divided into three stages: database acquisition, pattern learning, and testing. First, during the database acquisition stage, base-illumination images of 3D-printed objects are captured, and ground-truth surface normal maps are obtained using their corresponding 3D modeling files. Next, in the pattern learning stage, optimized patterns for high-quality surface normal reconstruction are learned using a real-world training dataset. A detailed explanation of this process is provided in Section 6, *Reconstruction*. Finally, during the testing phase, various real-world objects are photographed using the learned patterns, and their surface normals are reconstructed.

**3.2. Difference 1: Mobile Imaging System**
DDPS uses an LCD monitor and a polarized camera to achieve diffuse-specular separation and improve surface normal reconstruction by focusing exclusively on the diffuse image. Additionally, the LCD monitor provides high light intensity. In contrast, DMDPS uses a mobile phone for both pattern display and image capture. To achieve simultaneous pattern display and image capture, a custom application was developed. However, the phone's front camera cannot perform diffuse-specular separation, and the light intensity is relatively low when using a phone as the light source.

## 3.3. Difference 2: Capture and Processing

Traditional display photometric stereo methods, including DDPS, typically require a dark room for image capture. External light sources weaken the effect of the designated light patterns, leading to poor-quality surface normal reconstruction as the captured images look similar regardless of the pattern used.

DMDPS overcomes this limitation by enabling image capture in general environments. HDR imaging is employed to mitigate the impact of external lighting, maximizing the contrast between bright and dark areas and improving surface normal reconstruction quality.

# 4. Mobile Display System

## 4.1. Imaging System

We used a mobile phone for both pattern display and image capture. The device utilized was a Galaxy S22, equipped with a 6.1-inch display (19.5:9 aspect ratio) and a resolution of 2340 x 1080 pixels. The front camera is a 10-megapixel sensor. Image capture was performed in a standard room rather than a dark room. The mobile phone was mounted on a cradle, and the custom app was used for image capture.

Since the imaging device shifted from an LCD monitor to a mobile phone, camera calibration was necessary. Camera calibration determines the internal parameters required to convert image coordinates into world coordinates. The transformation formula is as follows:

$$\tilde{x} = \begin{bmatrix} f & 0 & p_x & 0 \\ 0 & f & p_y & 0 \\ 0 & 0 & 1 & 0 \end{bmatrix} \begin{bmatrix} R & -Rt \\ 0 & 1 \end{bmatrix} \widetilde{X_w},$$

where $\tilde{x}$ represents the homogeneous coordinate of the 2D image point, f denotes the focal length, and $p_x, p_y$ are the difference between the image coordinate and the camera coordinate. $R$ represents rotation, and $t$ denotes the translation between the world coordinate and the camera coordinate. $\widetilde{X_w}$ is the homogeneous coordinate of the 3D world point.

Camera calibration was performed using MATLAB. Images of a checkerboard taken at various angles were processed with MATLAB's camera calibration tool to determine the camera's internal parameters, including focal length and principal point.

## 4.2. Capture Application Design

When using a mobile phone instead of a polarized camera to capture images, we encountered the following challenges:

**Shared Device for Pattern Display and Image Capture**  The mobile phone must simultaneously display multiple patterns and capture images using its front camera while each pattern is reflected on the object.

**Raw Image Requirement**  General photography formats like JPG or PNG undergo processing steps such as noise reduction, gamma adjustment, and compression of highlights and shadows. These processes result in a loss of dynamic range and image information. To preserve the dynamic range needed for HDR imaging, RAW images were used instead of processed formats like JPG or PNG.

**Reduced Light Intensity**  The mobile phone's light source is significantly smaller and less intense than an LCD monitor. To compensate, the ISO value was increased to enhance light intensity. However, raising the ISO introduces noise. To address this, the ISO was set to 50 (the lowest possible value), and light intensity was increased by extending the exposure time.

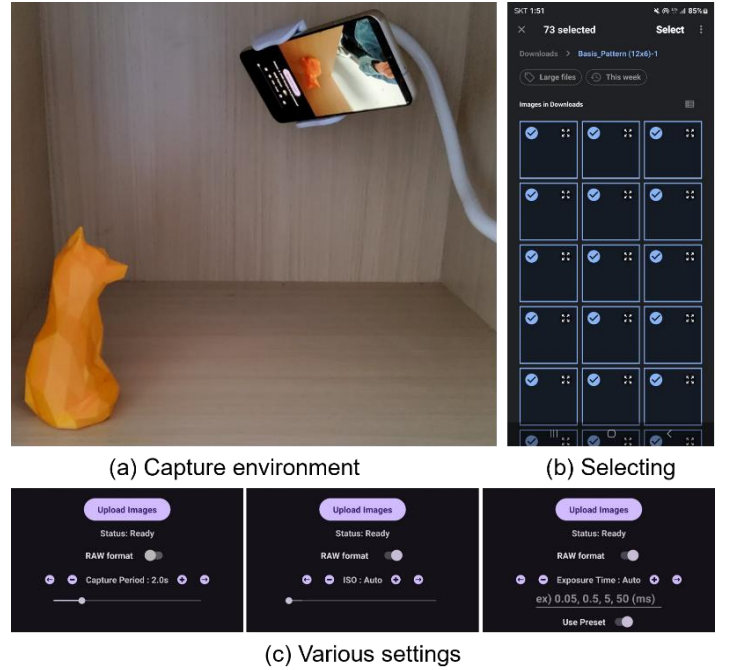

Figure 2. **Overview of the application.** (a) The app we developed displays how pictures are currently being taken using the front camera. Users can view the screen, position the object, and take a picture. (b) By pressing the "Upload Image" button in the app, users can upload the desired patterns for display. (c) The app allows users to adjust various settings, such as ISO and exposure time.

**Adjustable Exposure Time for Light Control**  The app was designed to allow adjustment of exposure time to control light intensity in captured images. Since increasing the ISO value was limited, exposure time was adjusted to brighten the image. Additionally, HDR imaging required capturing photos at multiple exposure times.

To address these challenges, we developed a new app. Figure 2 shows various screenshots of the app we developed. Figure 2(a) illustrates the shooting environment. The app displays the live feed from the mobile phone's front camera on its screen. Users can align the object with the mobile phone's position by referencing this live feed and then take a picture. Figure 2(b) demonstrates the pattern upload feature. Users can press the "Upload Images" button, select the desired pattern to display, and upload it. Simultaneously, the front camera of the mobile phone captures an image of the object. Figure 2(c) highlights the app's various settings. Users can capture RAW images by pressing the RAW format button. They can also adjust the ISO value or exposure time using a slide bar. Furthermore, the "Use Preset" option allows users to capture images at multiple preset exposure times. Figure 3 shows a picture of an object according to several exposure times and ISO values.

Additionally, a green dot briefly appears in the top-right corner of the screen at the start of photographing due to device-specific behavior. If the exposure time is prolonged, the green dot becomes more noticeable.

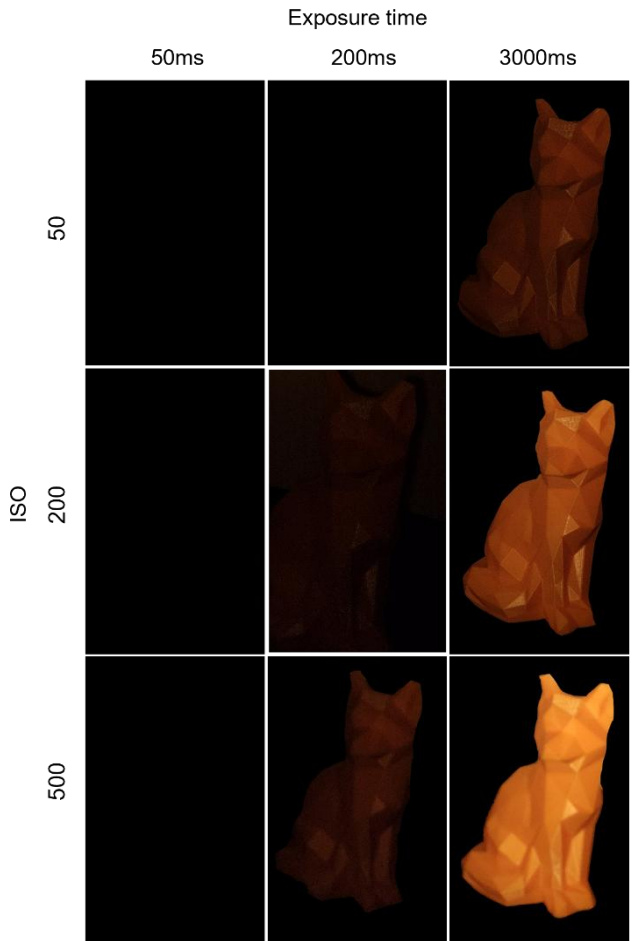

Figure 3. **Photos Captured with Different Settings.** These images show objects photographed with varying ISO and exposure time values.

To address this issue, a dummy image is captured before displaying the pattern, effectively reducing the green dot's impact on the final image.

### 4.3. Image Processing

To achieve high-quality surface normal reconstruction, followings were considered.

**Gaussian Filter**  Images captured with a mobile phone tend to exhibit more noise compared to those taken with an LCD monitor, making noise reduction essential. Various filters, including average filters and median filters, were tested. The Gaussian filter yielded the best results for surface normal reconstruction by assigning greater weight to pixels closer to the target point, thereby smoothing the image effectively.

**HDR Imaging**  Our study proposes capturing objects in general environments instead of a dark room. In such environments, external light sources—other than the intended illumination—can interfere with the image capture process. This interference reduces the influence of the displayed pattern's light, causing the captured images to appear similar regardless of the pattern used.

This issue is particularly problematic when capturing basis images [2]. Ideally, these images should vary based on the positions of several superpixels. However, strong external lighting and the limited brightness of mobile phone displays can diminish this variation, leading to poor reconstruction results.

To address this issue, we adopted high dynamic range (HDR) imaging. HDR enhances the dynamic range by maximizing the contrast between bright and dark areas in an image. HDR imaging involves two primary steps:

**Merging**  The captured low dynamic range (LDR) images are combined into a single HDR image. Each pixel is weighted during this process, as represented by the merging equation,

$$I_{HDR}(x,y) = \sum_{t_i} W(I_{t_i}(x,y)) \frac{I_{t_i}(x,y)}{t_i},$$

where W is weight function.

**Exposure Bracketing**  This process, discussed later in Section 4.5, involves capturing images at multiple exposure times to create HDR images.

**Undistortion**  Mobile cameras typically have a wide field of view (FOV), resulting in significant distortion. This distortion manifests as both radial distortion and tangential distortion. Radial distortion is caused by the lens's refractive index, with distortion intensity increasing as the distance from the image center grows. The distorted position is represented as follows:

$$x_{distorted} = x(1 + k_1 r^2 + k_2 r^4 + k_3 r^6),$$

where $x$ is x-coordinates in an undistorted image, $k_1, k_2, k_3$ is radial distortion factor and $r^2 = x^2 + y^2$.

Tangential Distortion caused by misalignments during assembly, such as a misaligned lens center or lack of horizontality between the lens and sensor. Distortion distribution varies in the form of an ellipse. The equation for the distorted position is as follows.

$$x_{distorted} = x(2p_1 y + p_2(r^2 + 2x^2))$$

To correct these distortions, distortion coefficients ($k_1,k_2,p_1,p_2,k_3$) were obtained through camera calibration. The undistortion process involves:

1. *Normalization*: Converting the captured image to normalized coordinates using the inverse matrix of the camera's internal parameters.

2. *Distortion Application*: Applying radial and tangential distortion models to the normalized coordinates.

3. *Undistortion*: Reapplying the camera's internal parameters to the distorted coordinates to correct the image.

Figure 4(a) shows the camera calibration process, and Figure 4(b) illustrates the superpixel calibration for the front camera. Undistortion significantly improved reconstruction quality. Figure 5 shows the difference between before and after the undistortion was applied. The difference between the ground truth and the estimated normal map decreased from 0.003662 to 0.003243 after applying undistortion.

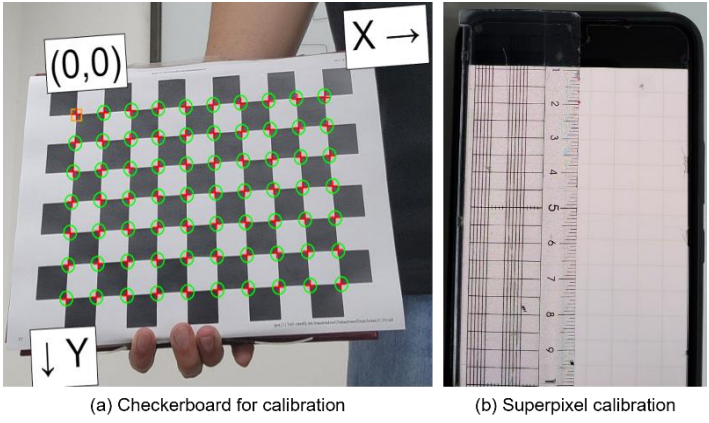

(a) Checkerboard for calibration  (b) Superpixel calibration

Figure 4. **Calibration.** (a) Several photos of a checkerboard pattern were captured using the target device, and camera calibration was performed. (b) The positions of individual superpixels relative to the front camera were calibrated using a grid plate and ruler.

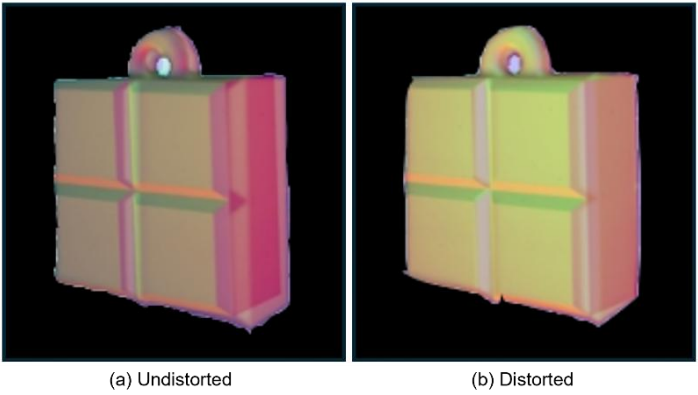

(a) Undistorted  (b) Distorted

Figure 5. **Distorted vs. Undistorted images.** (a) Shows a distorted image, while (b) shows the same image after undistortion has been applied.

### 4.4. Capture Process

The image capture process was carried out using the custom app. After setting the desired options, the phone was placed on a cradle. When a pre-downloaded pattern was uploaded using the image upload button, the app automatically displayed the patterns and captured images with the front camera. Additionally, exposure bracketing was performed for HDR image production.

**Exposure Bracketing**: This process involves capturing multiple low dynamic range (LDR) images at varying exposure times. The image formation process is represented as follows:

$$I_{t_i}(x, y) = clip[t_i \cdot I(x, y) + noise],$$

where $t_i$ denotes the exposure time. $I(x,y)$ represents the intensity at a specific pixel (x,y) in the captured LDR image.

Using preset settings, we captured images at various exposure times. The captured RAW images were then merged into HDR images through exposure bracketing. These HDR images were used for surface normal reconstruction, resulting in improved outcomes.

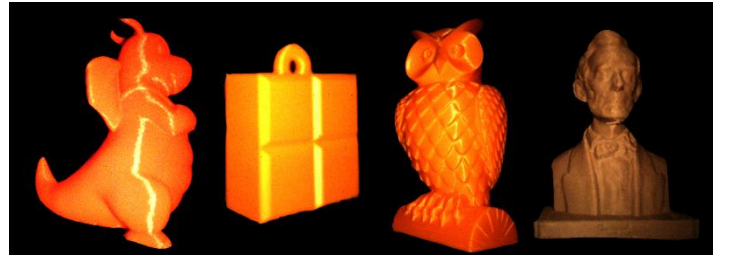

Figure 6. **Training sets.** Raw images from the training dataset captured using the DMDPS app.

| Illumination patterns | step_size ($\alpha = 5$) | step_size ($\alpha = 10$) | step_size ($\alpha = 15$) | step_size ($\alpha = 20$) |
|---|---|---|---|---|
| Mono-gradient | 0.06439 | 0.063 | 0.06189 | 0.063 |
| OLAT | 0.0684 | 0.07576 | 0.07749 | 0.07673 |
| Mono-random | 0.07387 | 0.06616 | 0.06567 | 0.06488 |
| Average | 0.06888 | 0.0683 | 0.06835 | 0.0682 |

Table 1. Reconstruction errors based on $\alpha$ and illumination patterns.

## 5. Reconstruction

### 5.1. Pattern Learning

To learn display patterns, we utilize a 3D-printed training dataset containing ground-truth normal maps $N_{GT}$ and a basis image **B**. Figure 6 shows a picture of training datasets. The K display patterns are denoted as **P**, which serve as our optimization variables.

We implemented an optimization pipeline using a differentiable image formation function $f_I$ and a differentiable photometric stereo method $f_n$. Function $f_I$ simulates captured images **I** for the training scene based on the display patterns being optimized. Subsequently, $f_n$ processes these simulated images to estimate surface normal **N**.

For a given display pattern $P_i$ and basis image **B**, the raw image simulation is expressed as:

$$I_i = f_I(P_i, \mathbf{B}) = \sum_{j=1}^{b} B_j P_{i,j},$$

where $\boldsymbol{P}_{i,j}$ is the RGB intensity of the j-th superpixel in $\boldsymbol{P}_i$. This process is performed on K patterns to obtain set of synthesized images **I**.

$$\underset{\boldsymbol{P}}{minimize} \sum_{\boldsymbol{B},\boldsymbol{N_{GT}}} loss(f_n(\{f_I(P_i, \mathbf{B})\}_{i=1}^K, \boldsymbol{P}), \boldsymbol{N_{GT}}),$$

where the loss function penalizes the angular difference between estimated and ground-truth normal:

$$loss = (1 - N \cdot \boldsymbol{N_{GT}})/2$$

To ensure the display patterns remain within a physically valid intensity range [0,1], a sigmoid function is applied to **P** during optimization. The Adam optimizer is used to minimize this loss [11].

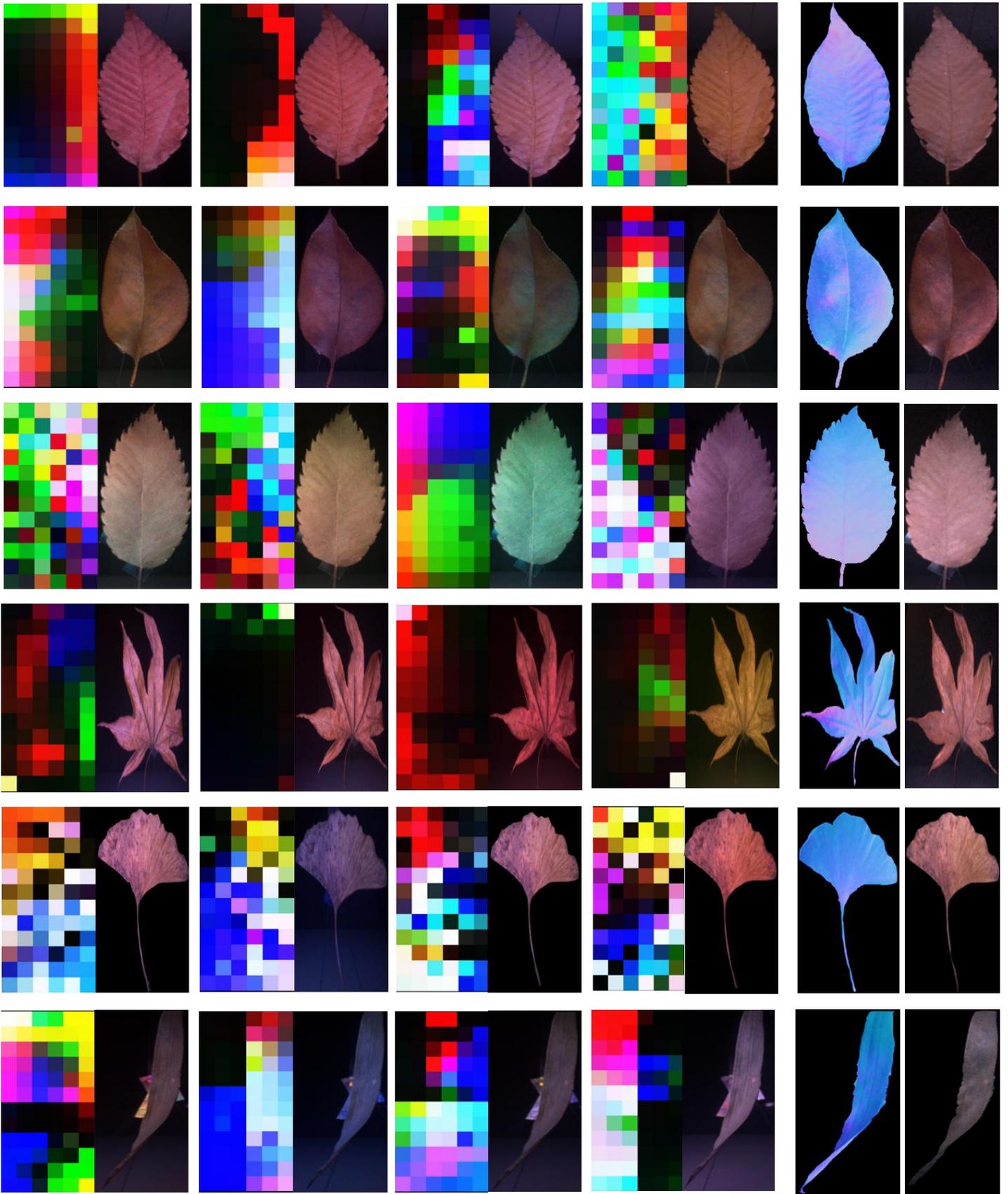

(a) Learned patterns and captured scenes

(b) Reconstructed normals & albedo

Figure 7. **Reconstructed results.** Surface normal reconstruction of various leaves using the learned patterns with DMDPS.

| Illumination Patterns | Number of patterns | Reconstruction Error | |
|---|---|---|---|
| | | Initial | DMDPS |
| OLAT | 4 | 0.1783 | 0.05326 |
| Group OLAT | 4 | 0.0912 | 0.05265 |
| Mono-gradient | 4 | 0.0937 | 0.05174 |
| Mono-random | 4 | 0.2795 | 0.05013 |
| Tri-gradient | 2 | 0.0984 | 0.05668 |
| Tri-random | 4 | 0.1573 | 0.05834 |
| Flat gray | 4 | 0.4109 | 0.05754 |
| Mono-complementary | 4 | 0.1090 | 0.05087 |
| Tri- complementary | 2 | 0.1013 | 0.06918 |

Table 2. Reconstruction error for each initial pattern.

The learning rate for pattern learning was adjusted indirectly by modifying the step size ($\alpha$) variable in the code rather than altering the learning rate value directly. $\alpha$ controls the application of a decay factor (set to 0.3) to the learning rate at each epoch. The performance of surface normal reconstruction was evaluated for various $\alpha$ values and illumination patterns. Table 1 presents the reconstruction errors under these configurations.

### 5.2. Normal Reconstruction

Surface normal reconstruction is performed using images captured or simulated under the optimized display patterns $P$. For a given pixel, the RGB intensity under the i-th display pattern is denoted as $I_i^c$, where $c \in \{R, G, B\}$. Illumination from the j-th superpixel is represented by the spatially varying vector $l_j$, computed based on the relative position of the superpixel and the scene point. Assuming a planar surface at a fixed distance of 10 cm from the camera, the relationship between observed intensities and surface normals is modeled as:

$$\mathbf{I} = \boldsymbol{\rho} \odot \boldsymbol{P}l\boldsymbol{N},$$

where $I \in \mathbb{R}^{3K \times 1}$ is vectorized intensity, $\rho \in \mathbb{R}^{3K \times 1}$ is albedo, $N \in \mathbb{R}^{3 \times 1}$ is surface normal, $P \in \mathbb{R}^{3K \times b}$ is pattern intensity matrix, $l \in \mathbb{R}^{b \times 3}$ is illumination direction matrix and $\odot$ is Handamard product.

The unknowns in this equation are the surface normals $N$ and albedo $\rho$. For numerical stability, $\rho$ is set as the maximum intensity observed across captures. The surface normals are computed using the pseudo-inverse method:

$$N \leftarrow (\boldsymbol{\rho} \odot \boldsymbol{P}l)^{\dagger} \mathbf{I},$$

where † denotes the pseudo-inverse operator. This reconstruction method avoids the need for trainable parameters, relying instead on analytic formulations. Once the display patterns are optimized, they are tested on real-world objects by capturing images under these patterns and reconstructing surface normals using the photometric stereo method ($N = f_n(I)$).

## 6. Assessment

We assess DMDPS on diverse objects.

### 6.1. Learned Patterns & Reconstruction Error
Figure 7 shows the learned patterns, photos of objects taken with those patterns, and the surface normals and albedos restored with DMDPS. Figure 8 shows what the learned patterns look like for each initial pattern. Table 2 summarizes the reconstruction errors of each pattern for the 3D-printed database.

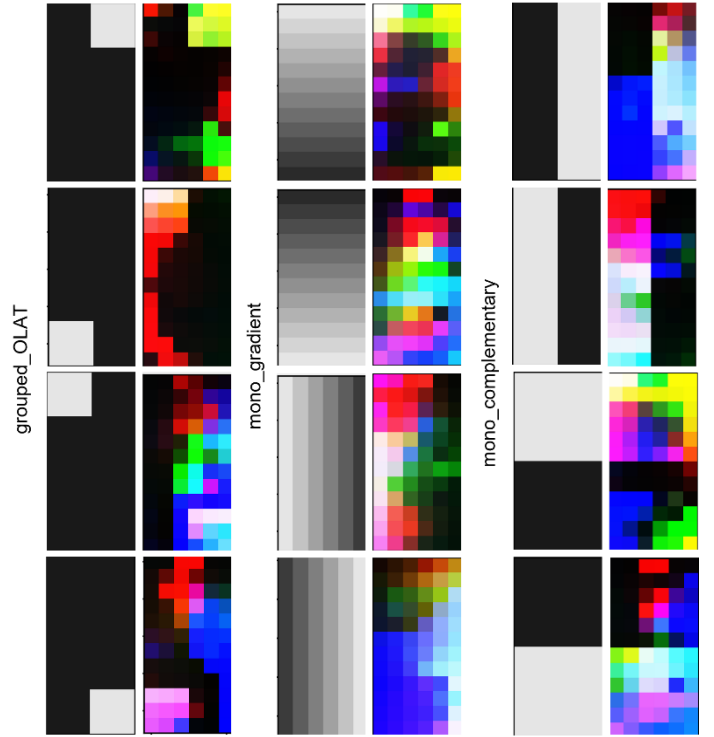

Figure 8. **Initial & Learned patterns.** The left side of each row represents the initial patterns, while the right side shows the learned patterns.

On average, the reconstruction error across all patterns was approximately 0.01 higher than that of DDPS. This increase can be attributed to the following factors:

· *External Light Interference*: Although HDR imaging partially mitigated the impact of external light, the intensity of external light relative to the phone's illumination remained significant, contributing to residual errors.

· *Diffuse-specular Separation*: DDPS employed a polarization camera and LCD monitor to sperate diffuse. In contrast, some mobile phone models do not emit polarized light, and the front camera is a standard, non- polarization camera. Consequently, DMDPS used both diffuse and specular images without separation, which likely increased reconstruction errors.

### 6.2. Glossy Object
While our DMDPS method demonstrated strong performance across a wide range of objects, glossy objects posed notable challenges, resulting in significantly lower-quality surface normal reconstructions. Glossy surfaces, such as those found on metallic or highly reflective objects like metal cans, introduce complexities due to their strong specular reflections.

Figure 9 illustrates an example of this phenomenon, where the surface normal reconstruction for a glossy object was substantially degraded. The primary issue lies in the interaction of light with highly reflective surfaces. Glossy objects exhibit pronounced specular reflections that dominate the captured intensity and obscure the diffuse components necessary for accurate photometric stereo reconstruction.

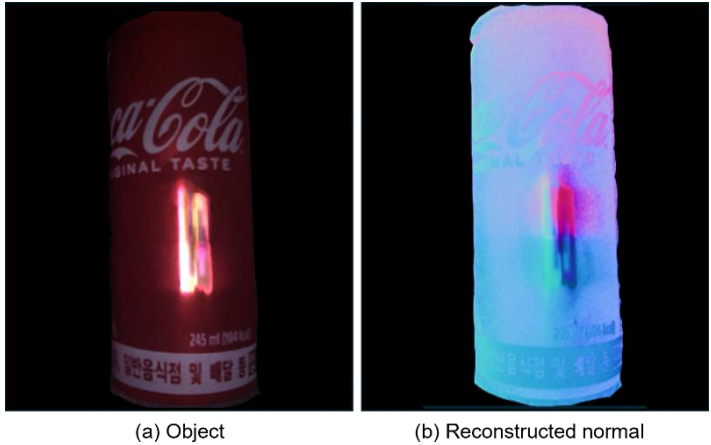

Figure 9. **Glossy material (can).** (a) A glossy object (can) and (b) its reconstructed surface normal. The results show that the quality of the surface normal reconstruction is poor for glossy materials.

Unlike matte surfaces, glossy surfaces reflect incoming light directionally, creating highlight regions that are highly sensitive to viewing and lighting angles. These reflections lead to non-linearities in the captured images.

Additionally, glossy surfaces often cause environmental light and unintended reflections to interfere with the captured images, introducing further artifacts and noise. For instance, a glossy metal can might reflect not only the light from the display patterns but also other elements in the surrounding environment, such as the camera lens or nearby objects.

In summary, while DMDPS performs robustly across most object types, glossy objects remain a challenging edge case due to the dominance of specular reflections and environmental interference. Addressing these limitations will be crucial to extending the applicability of our method to real-world scenarios involving highly reflective materials.

### 6.3. Natural Object

After validating our method on 3D-printed objects, we extended the evaluation to more diverse and unstructured natural objects to assess the robustness and applicability of the approach in real-world scenarios. Among various candidates, we selected fallen leaves as the primary subject due to their intricate structures and diverse surface textures.

To systematically evaluate performance, we collected a diverse set of fallen leaves, varying in size, shape, and texture. A custom dataset was created specifically for the surface normals of fallen leaves, which involved capturing multiple images of each leaf under the optimized display patterns. For each leaf, its position was carefully controlled to maintain a consistent distance of 10 cm from the camera.

Figure 7 illustrates the reconstructed surface normals for several types of fallen leaves. The results highlight the capability of our method to capture fine details of leaf textures, such as veins and subtle undulations, as well as handle variations in curvature. The high angular accuracy of the reconstructed normals demonstrates that the learned display patterns are robust, even for complex, non-planar surfaces.

This evaluation on natural objects not only validated the versatility of our approach but also highlighted potential areas for further improvement. By expanding the application to fallen leaves, we demonstrated the feasibility of applying our framework to natural, unstructured environments, paving the way for broader applications.

## 7. Discussion

First, we were unable to perform diffuse-specular separation due to device limitations, which we believe is a significant factor to the reconstruction error. For future work, we propose exploring whether diffuse-specular separation can be achieved on mobile devices by using a model capable of emitting polarized light, coupled with a polarization lens or polarizing film attached to the front camera.

Second, although our goal was to create a large dataset of normal maps by photographing fallen leaves of various shapes, we were able to capture only a subset of the fallen leaves. Expanding the dataset by photographing a wider variety of fallen leaves remains a future task.

Lastly, it may be interesting to explore the use of tablets instead of mobile phones. Tablets offer the portability of a mobile phone while featuring a larger screen, which can serve as a larger light source. These characteristics suggest the potential for producing high-quality surface normal reconstructions, making the application of DDPS to tablets a promising area for future research.

## 8. Conclusion

In this paper, we presented the DMDPS. Surface normal reconstruction was performed using a mobile phone instead of a desktop setup of DDPS. Monitor and polarization camera, and objects were photographed in general environments rather than in a dark room. Tests were conducted using 3D-printed objects, and normal maps were established, along with datasets created for natural objects (e.g., fallen leaves). We hope our work spur further interests in practical physics-based photometric stereo.